\documentclass[letterpaper, 10 pt, conference]{ieeeconf}  

\IEEEoverridecommandlockouts                              

\usepackage{graphicx}
\usepackage{cite}
\usepackage{amsmath,amssymb,amsfonts}
\usepackage{algorithm}
\usepackage{algorithmic}
\usepackage{comment}
\usepackage{caption}
\usepackage{subcaption}
\usepackage{multirow}
\usepackage{float}    
\usepackage{placeins} 
\usepackage{hyperref}
\usepackage{booktabs}
\usepackage[table,xcdraw]{xcolor}
\usepackage{svg}
\usepackage{tikz}
\usetikzlibrary{shapes.geometric, arrows, decorations.pathreplacing, positioning}
\usepackage{dblfloatfix}
\usepackage{eso-pic}
\newcommand{\mycopyrighttext}{%
  \footnotesize
  \noindent
  \textcopyright~2025 IEEE. Personal use of this material is permitted. Permission from IEEE must be obtained for all other uses, in any current or future media, including reprinting/republishing this material for advertising or promotional purposes, creating new collective works, for resale or redistribution to servers or lists, or reuse of any copyrighted component of this work in other works.\\
 IEEE/RSJ International Conference on Intelligent Robots and Systems (IROS) - October 19-25, 2025.
}

\AtBeginDocument{%
  \AddToShipoutPicture*{%
    \AtPageUpperLeft{%
      \put(55, -40){
        \parbox{\dimexpr\textwidth-4pt\relax}{\raggedright \mycopyrighttext}
      }%
    }
  }
}

\title{\LARGE \bf Diverse and Adaptive Behavior Curriculum for Autonomous Driving: \\ A Student-Teacher Framework with Multi-Agent RL}
\author{Ahmed Abouelazm\textsuperscript{\textasteriskcentered}$^{1}$, Johannes Ratz\textsuperscript{\textasteriskcentered}$^{2}$, Philip Schoerner$^{1}$, and J. Marius Zöllner$^{1,2}$
\thanks{\textasteriskcentered~These authors contributed equally to this work}%
\thanks{$^{1}$Authors are with the FZI Research Center for Information Technology, Germany
        {\tt\small abouelazm@fzi.de}}%
\thanks{$^{2}$Authors are with the Karlsruhe Institute of Technology, Germany}%
}

\begin{document}
\maketitle
\thispagestyle{empty}
\pagestyle{empty}

\begin{abstract}
    Autonomous driving faces challenges in navigating complex real-world traffic, requiring safe handling of both common and critical scenarios. 
Reinforcement learning (RL), a prominent method in end-to-end driving, enables agents to learn through trial and error in simulation. However, RL training often relies on rule-based traffic scenarios, limiting generalization. Additionally, current scenario generation methods focus heavily on critical scenarios, neglecting a balance with routine driving behaviors. Curriculum learning, which progressively trains agents on increasingly complex tasks, is a promising approach to improving the robustness and coverage of RL driving policies. However, existing research mainly emphasizes manually designed curricula, focusing on scenery and actor placement rather than traffic behavior dynamics. This work introduces a novel student-teacher framework for automatic curriculum learning. The teacher, a graph-based multi-agent RL component, adaptively generates traffic behaviors across diverse difficulty levels. An adaptive mechanism adjusts task difficulty based on student performance, ensuring exposure to behaviors ranging from common to critical. The student, though exchangeable, is realized as a deep RL agent with partial observability, reflecting real-world perception constraints. Results demonstrate the teacher’s ability to generate diverse traffic behaviors. The student, trained with automatic curricula, outperformed agents trained on rule-based traffic, achieving higher rewards and exhibiting balanced, assertive driving.
    
\end{abstract}
\section{Introduction}
\label{sec:Introduction}
Car accidents cause 1.19 million deaths annually, with road injuries the leading cause of death for ages 5–29~\cite{whoRoadTrafficInjuries}
, prompting research into safe and reliable autonomous vehicles. 
Traditional autonomous systems employ a modular structure with distinct modules for tasks like perception, localization, and planning. However, this approach often leads to error propagation and requires significant engineering effort. Additionally, as each module is optimized independently, the overall system may lack alignment toward a unified goal, such as ensuring safe and efficient driving~\cite{hu2023planning}. End-to-end (E2E) driving systems aim to address these challenges by enabling joint learning across tasks, often mapping raw sensor data to control actions or intermediate representations. These systems enable joint optimization of perception, prediction, and planning, making them well-suited for complex scenarios and reacting to hazards~\cite{chen2023endtoend}. Reinforcement learning (RL) has been explored as an E2E approach~\cite{al2024autonomous}, enabling systems to learn optimal driving policies through trial and error. 
Agents receive rewards based on task performance, such as maintaining lane position or avoiding collisions, allowing continuous refinement of their driving policy. RL agents are typically trained in simulations, enabling safe exploration without real-world risks~\cite{chen2023endtoend}. 

\textbf{Research Gap. }RL agents learn through interaction with the environment, enabling them to handle traffic participant behaviors similar to those encountered during training. Simulation environments must encompass a wide range of diverse, common, and safety-critical behaviors to achieve robustness and generalization. However, most simulations rely on fixed non-player characters (NPCs) behaviors with predefined rules, maintaining constant speed or distance~\cite{dosovitskiy2017carla}. This limits the agent's ability to generalize to unseen situations~\cite{wang_advsim_2023}. Similarly, prerecorded human driving behaviors~\cite{caesar2022nuplan} offer realism but are non-adaptive and rarely include safety-critical situations, contributing to the \textit{long tail problem}~\cite{hanselmann2022king}, where critical events are underrepresented.

Recent work targets safety-critical behaviors by generating scenarios that provoke the self-driving vehicle (SDV) collisions or off-road events~\cite{hanselmann2022king, liu_safety-critical_2023}. While effective for robustness against safety-critical situations, they neglect more frequent traffic patterns, potentially limiting generalization to everyday driving contexts. To overcome this limitation, more generalizable approaches such as multi-agent reinforcement learning have been explored. Multi-agent RL (MARL) methods~\cite{peng2021learning} allow NPCs to make independent or cooperative decisions but primarily focus on traffic coordination rather than training SDV agents. To develop driving skills more comprehensively, some works adopt curriculum learning~\cite{bengio2009curriculum}, a technique inspired by education that trains agents on simpler tasks before progressively advancing to more complex ones. This approach improves learning efficiency and robustness. However, in autonomous driving, current research relies on manually crafted task sequences~\cite{anzalone2022end}, which fail to encompass a diverse range of NPC behaviors. 
\textbf{Contribution.}
This work introduces a student-teacher framework for adaptively generating a behavior curriculum for surrounding NPCs that dynamically adapts to the student's performance. By integrating MARL and curriculum learning, the framework enhances the generalization and robustness of the student, enabling it to handle a diverse spectrum of traffic behaviors. The key contributions of this work are:
\begin{itemize}
    \item \textbf{Teacher Design}: A novel MARL-based teacher capable of generating traffic behaviors with varying difficulty levels with a graph-based network and a novel reward.
    
    \item \textbf{Automatic Curriculum Algorithm}: A method for automatically orchestrating the concurrent training of student and teacher components, creating an adaptive behavior curriculum.
    
\end{itemize}

\section{related work}
\label{sec:sota}
This section reviews prior research on traffic simulation, with an emphasis on behavior generation techniques. Additionally, curriculum learning is discussed, highlighting advancements in automatic curriculum generation.
\begin{figure}[t]
    \centering
    \includegraphics[trim={9.5cm 2cm 9.5cm 1.4cm},clip,width=0.9\columnwidth]{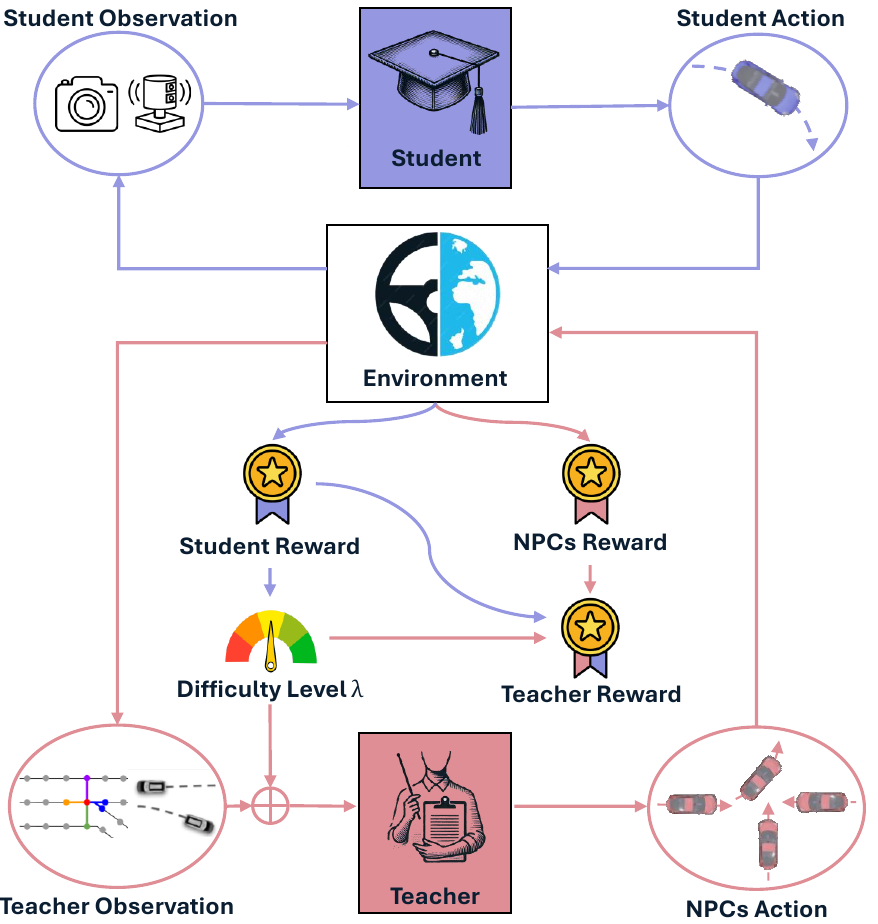}
    \caption{Overview of the proposed behavior curriculum framework. The student (blue) and teacher (red) interact concurrently in the driving environment. The student learns a driving policy from sensor data, guided by a standard reward. The teacher dynamically orchestrates NPC behavior based on a fully observable state representation, the student's performance, and an auxiliary input $\lambda$. 
    }
    \label{fig:concept}
    \vspace{-0.6cm}
\end{figure}
\subsection{Scenario Generation}
Traditional Simulators like CARLA~\cite{dosovitskiy2017carla} 
generate rule-based NPC behavior, ensuring compliance with traffic laws, lane-keeping, and collision avoidance. 
However, they lack flexibility and require manual adjustments to parameters such as speed limits. In addition, they fail to simulate unpredictable behavior or dynamically adapt to agents' actions. 

TrafficGen~\cite{feng2023trafficgen} addresses these limitations with a generative model trained on real-world scenarios to produce realistic vehicle placements and trajectories. It generates scenarios autoregressively, predicting long-term motion. However, it relies heavily on large datasets, struggles with out-of-distribution road topologies, and lacks adaptability to agent actions. TrafficSim~\cite{suo2021trafficsim} enhances the learning objective with a \textit{common-sense} loss and predicts multi-modal trajectories to capture the uncertainty in agent interactions. 
However, the inclusion of a common-sense loss requires a differentiable simulator, making the approach computationally intensive.

CoPO~\cite{peng2021learning} eliminates the need for differentiable simulators by introducing a MARL algorithm that coordinates traffic flow, allowing NPCs to exhibit cooperative and competitive behaviors. 
While the aforementioned approaches effectively generate a wide range of traffic behaviors, they are not tailored to capture safety-critical situations. In contrast, AdvSim~\cite{wang_advsim_2023} generates safety-critical scenarios by modifying real-world traffic through physically plausible perturbations of agents' trajectories. 
It employs black-box optimization (BBO)~\cite{alzantot2019genattack} to identify failure-inducing perturbations. However, BBO is computationally intensive in high-dimensional spaces~\cite{hanselmann2022king}. 
KING~\cite{hanselmann2022king} employs a gradient-based procedure to optimize trajectory perturbations iteratively, eliminating the need for BBO. However, this approach requires differentiable approximations of simulator aspects, such as vehicle dynamics, and loss components, such as collision.

Liu et al.~\cite{liu_safety-critical_2023} use RL to edit scenarios by adding agents or perturbing their trajectories. 
However, pre-setting trajectories before scenarios unfold limits their ability to capture the impact of agents' behaviors and interactions at specific time steps on the SDV. 
Conversely, FailMaker-AdvRL~\cite{wachi_failure-scenario_2019} leverages MARL to coordinate adversarial agents, introducing techniques like \textit{Contributors Identification} to rank agents' contributions to SDV failures and \textit{Adversarial Reward Allocation} to assign reward fractions based on their impact on the failure. These approaches often rely on scene-specific perturbations and fail to learn a generalizable policy. 

\subsection{Curriculum Learning}
Instead of explicitly training SDV agents on specific traffic behaviors, some methods adopt curriculum learning to generate scenarios of increasing difficulty. This approach aims to shorten training time and enhance robustness across a range of tasks. Approaches proposed in~\cite{anzalone2022end, qiao2018automatically} manually design multi-stage curricula by increasing the number of agents, but overlook variations in NPCs' behavior. Khaitan and Dolan~\cite{khaitan2022state} introduce a state dropout-based curriculum, progressively removing privileged information about surrounding vehicles' future states. However, these methods rely heavily on manual effort and expert knowledge.

To address the challenges of manual curriculum design, automated curriculum learning approaches have been introduced. 
A framework for Unsupervised Environment Design (UED)~\cite{dennis2020emergent} enables automatic modification of environments (or tasks) to optimize specific objectives. PLR~\cite{jiang2021replay} extends UED by combining curriculum learning with evolutionary methods to generate increasingly complex tasks, using regret—which measures the gap between the agent’s performance and an optimal policy—as a fitness objective. More advanced methods like ACCEL~\cite{parker2022evolving} further improve sample efficiency and refine promising tasks.

ARLPCG~\cite{gisslen2021adversarial} employs a dynamic task generation framework with a solver-generator setup (Solver and Generator). Unlike ACCEL, it adopts an adversarial approach where the two RL agents operate simultaneously in the same environment. The Generator is conditioned on an auxiliary input to control task difficulty, enhancing the Solver's generalization through exposure to diverse scenarios.

A key challenge in automatic curriculum learning, and a motivation for this paper, is the tendency to treat tasks or environments as a whole, overlooking agent behavior at individual simulation steps. Existing methods focus on layout and actor placement, neglect NPC reactive behavior, and are limited to benchmark games with no clear path to autonomous driving applications.
\section{Methodology}
To address the limitations of traffic generation approaches outlined in Section~\ref{sec:sota}, we present an automated curriculum framework for behavior generation, as illustrated in Fig.~\ref{fig:concept}. The framework adopts a student-teacher paradigm, where the student represents the ego vehicle, learning to navigate safely toward a destination. On the other hand, the teacher’s primary role is to dynamically adapt NPC behaviors based on a desired difficulty level, referred to as auxiliary input \(\lambda\)~\cite{gisslen2021adversarial} and feedback on the student’s performance, based on its driving reward. These components are combined into a teacher reward signal that guides the teacher in generating traffic behaviors suited to the current difficulty. This creates a symbiotic relationship between teacher and student, where behaviors are continuously adapted to the student’s capabilities, fostering structured and effective learning. 
The following subsections provide a detailed discussion of the framework components and the underlying algorithm.

\subsection{Teacher}
\label{subsection:teacher}
\begin{figure*}[t]
    \centering
    \includegraphics[trim={1.5cm 6.5cm 1.5cm 6.5cm},clip,width=\textwidth]{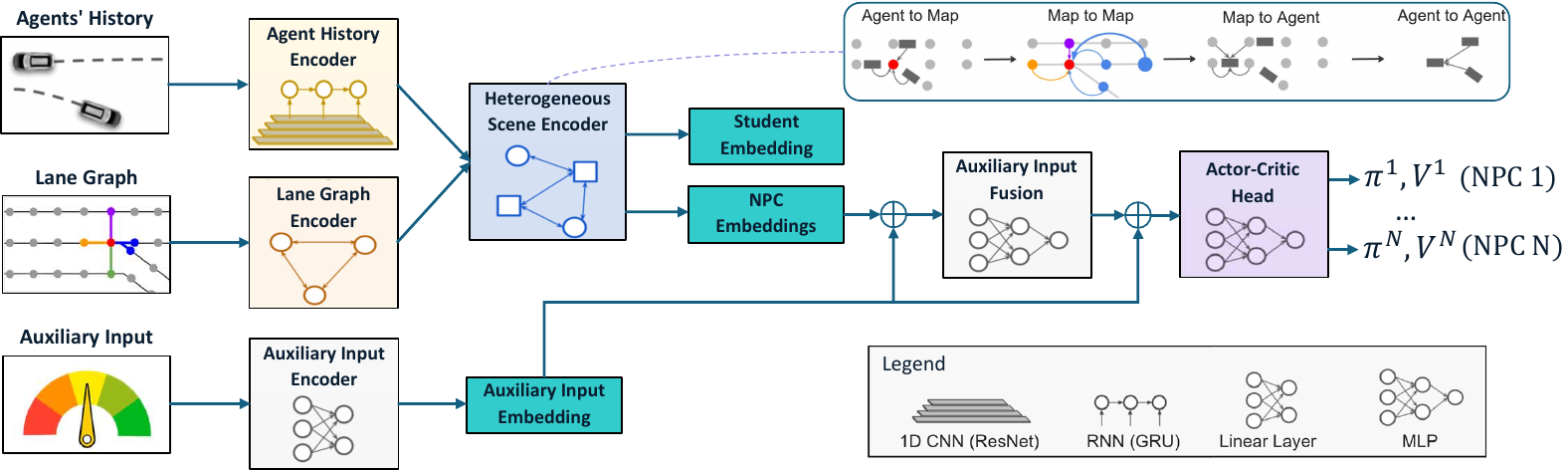}
    \caption{The teacher network architecture first encodes the agents' history and map lane graph separately, followed by hierarchical interaction fusion, ensuring each NPC embedding incorporates information about the map topology, road layout, surrounding NPCs, and the student. The NPC embeddings are concatenated and combined with a linear projection of the auxiliary input before being processed by the actor-critic MLP, which outputs the policies and value functions for each NPC.}
    \label{fig:network}
    \vspace{-0.6cm}
\end{figure*}
The teacher's role is to coordinate NPC behavior to generate scenarios matching a desired difficulty level $\lambda$. Unlike critical scenario generation frameworks, our approach aims to train the teacher to learn a generalizable policy applicable across diverse scenarios rather than being overfit to an individual one. Imitation learning is inadequate for our framework as it relies on expert trajectories, which are unavailable for the diverse range of behaviors and difficulties we aim to capture in our curriculum. Instead, we leverage a MARL-based teacher, which is effective in training policies for multi-agent systems~\cite{wong2023deep}, with each NPC treated as an individual agent. In the following, we discuss key aspects of the teacher, including its observation space, the reward function that drives its training, and the algorithm employed.

\subsubsection{Observation Space}
To effectively coordinate traffic, the teacher’s observation space is designed to be fully observable. This allows the teacher to access the complete state of the environment, providing the necessary context to manage complex scenarios. Unlike SDVs, which are limited by sensor-based observations, the teacher leverages its full observability to generate diverse and adaptive traffic behaviors in simulation. This paper introduces an observation space designed to encompass the historical motion of all NPCs and the student vehicle, along with a vectorized representation of road topology. This design draws inspiration from well-established input spaces used in trajectory prediction~\cite{liang2020learning}. The motion history of each agent is represented by a temporal sequence of poses. Each pose includes the agent's position, heading, 2D velocity, and acceleration, capturing all relevant information for modeling agent motion. Additionally, each agent is assigned a road option, which serves as a high-level navigational command (e.g., straight, turn left, or right) to guide NPCs toward their destinations~\cite{chitta2022transfuser}.

The road topology is modeled as a vectorized lane graph~\cite{liang2020learning}, where nodes represent lane segments and edges capture spatial and topological relationships. Node attributes include position, lane width, and curvature, as well as its type—e.g., crosswalk or intersection. Nodes are connected through various relations, such as predecessors, successors, left neighbors, and right neighbors. To enrich the graph's relational context, the predecessor and successor relationships are dilated with a dilation power \(D\). For example, when \(D = 1\), the graph includes not only direct successors but also the successor of a successor, providing additional contextual information. Finally, the auxiliary input \(\lambda\) is incorporated into the observation space to represent the desired traffic difficulty, ranging from easy (\(\lambda = 1\)) to most difficult (\(\lambda = -1\)). It is sampled from a discrete distribution, i.e., a predefined set of values within \([-1,1]\). Although $\lambda$ can be sampled from a continuous distribution, this would lead to longer training times and make it more challenging to design a curriculum with consistently difficult levels~\cite{gisslen2021adversarial}.

\subsubsection{Network Architecture}
The teacher employs a custom network architecture, as shown in Fig.~\ref{fig:network}, inspired by GoRela~\cite{cui2023gorela} to process high-dimensional observations efficiently. Unlike previous MARL behavior generation approaches that handle state representations for each NPC independently~\cite{peng2021learning}, our proposed network adopts a graph-based architecture that enables joint processing of a shared state representation of agents and the map lane graph. By leveraging a shared optimization process, the network effectively captures the relationships among a variable number of NPCs while preserving the independence of agents in decision-making, thereby enhancing behavioral diversity.

First, the map and agent representations are encoded using \textit{pair-wise relative positional encoding}, a viewpoint-invariant encoding approach~\cite{cui2023gorela}. Viewpoint invariance enhances the network’s ability to interpret the scene without being affected by rotation or translation of the scene viewpoint, thereby improving the teacher's robustness and eliminating the need for training data augmentation.

The agent-history encoder extracts feature embeddings from each agent's motion history using a 1D convolution layer with residual connections, effectively capturing both short- and long-term temporal dependencies. These extracted features are subsequently aggregated by a Gated Recurrent Unit (GRU)~\cite{cho2014learning}, where the final hidden state of the GRU serves as a compact embedding of the agent’s motion history. To incorporate the agent’s intended motion direction, we encode its road option using a multi-layer perceptron (MLP) and concatenate the resulting embedding with the history embedding. The map topology, represented as a lane graph, is processed through a lane graph encoder employing a stack of \textit{heterogeneous message passing} (HMP)~\cite{cui2023gorela} layers. HMP is particularly advantageous as it enables each lane segment’s embedding to be updated based on information from its neighboring segments connected through different types of spatial relationships. This allows the network to effectively capture both topological connectivity and the diverse geometric relationships among lane segments. Moreover, since the map topology remains fixed throughout a single driving scenario, the embeddings of map nodes are computed once per scenario and cached in memory, enhancing the network's computational efficiency.

Using the heterogeneous scene encoder, the network captures interactions between agents’ motion histories and map nodes embeddings. Unlike the heterogeneous scene encoder in GoRela, which processes all interactions in a single forward pass, our approach decomposes the interaction encoding into four sequential HMP layers, each dedicated to a specific type of relation: agent-to-map relations, which associate each agent with its neighboring lane segments; map-to-map relations, which update map node embeddings based on neighboring nodes; map-to-agent relations, which enable agents to receive spatial context from nearby map nodes; and agent-to-agent relations, which allow agents in proximity to exchange trajectory information and adjust their movements accordingly. 

The final agent embedding consists of two components: the student embedding, representing the ego vehicle, and the NPCs embedding, which encapsulates the surrounding NPCs whose motion is coordinated by the teacher. This sequential design ensures that the NPCs' embedding incorporates motion history, spatial context, and interaction-based information while remaining aware of the student's movement and position within the environment, resulting in a holistic representation of the driving scenario.

The auxiliary input is incorporated into the network through a linear layer and fused with the NPCs' embedding via concatenation and a fusion MLP, ensuring that the specified difficulty level is effectively embedded in NPC behavior. The resulting representation is then concatenated again with the auxiliary input embedding, reinforcing the network’s ability to account for the desired difficulty. Finally, an actor-critic MLP generates policies and value functions for each NPC based on its respective embedding. Thus, the proposed architecture enables the teacher to efficiently coordinate the behavior of a varying number of NPCs across different road topologies and difficulty levels.

\subsubsection{Reward Function}

In MARL, reward design plays a crucial role in shaping agents' interactions~\cite{du2019liir}, ensuring that cooperation, competition, and other dynamics align with the framework’s objectives. We propose a teacher reward that balances two competing objectives for each NPC: intrinsic and extrinsic rewards, as defined in Eq.~\ref{eq:reward}. The intrinsic reward, formulated in Eq.~\ref{eq:reward_int}, encourages realistic traffic behavior by promoting desirable driving traits such as goal-directed progress, collision avoidance, lane keeping, and comfort~\cite{reward_review}. Meanwhile, the extrinsic reward, defined in Eq.~\ref{eq:reward_ext}, provides feedback based on the student’s driving reward, enabling the teacher to adapt the NPC behavior to the student’s learning progress.
\begin{align}
    &{\mathcal{R}}_{\text{NPC}} = {\mathcal{R}}^{\text{intrinsic}}_{\text{NPC}} + {\mathcal{R}}^{\text{extrinsic}}_{\text{NPC}}\label{eq:reward}\\[3pt]
    {\mathcal{R}}^{\text{intrinsic}}_{\text{NPC}} &= (1 - K(d)) \cdot \max(\varepsilon, 1 - |\lambda|) \cdot {\mathcal{R}}^{\text{driving}}_{\text{NPC}}\label{eq:reward_int}\\[3pt]
    {\mathcal{R}}^{\text{extrinsic}}_{\text{NPC}} &= K(d) \cdot {\mathcal{R}}^{\text{driving}}_{\text{student}}\cdot 
    \begin{cases} \lambda & ,\text{if } |\lambda| > \varepsilon 
    \\ \text{sgn}(\lambda) \cdot \varepsilon &, \text{else} \end{cases}\label{eq:reward_ext}
\end{align}
The auxiliary input, \(\lambda \in [-1,1]\), serves as the primary balancing parameter, incentivizing NPCs to generate behaviors that align with the desired difficulty level. The intrinsic reward is weighted by \(1 - |\lambda|\), while the extrinsic reward is weighted by \(\lambda\), with both weighting terms clipped at $\varepsilon \in (0, 0.1]$ to prevent either reward objective from being entirely disregarded at any curriculum step. This weighting scheme enables the teacher to regulate NPC behavior based on the student's driving reward. At the easiest difficulty level (\(\lambda = 1\)), NPCs are highly altruistic, prioritizing the student's success and actively assisting in their progress. As \(\lambda\) decreases, NPCs become progressively more self-centered, prioritizing their intrinsic reward while reducing cooperation with the student, thereby increasing the scenario difficulty. When \(\lambda = 0\), NPCs exhibit predominantly egoistic behavior, optimizing their own objectives with minimal regard for the student's performance. As the difficulty increases (\(\lambda < 0\)), NPCs adopt an increasingly adversarial role, coordinating to hinder the student's progress, reaching its peak at \(\lambda = -1\). 

Additionally, we introduce a distance-based weighting mechanism to balance intrinsic and extrinsic reward objectives, employing a Radial Basis Function (RBF) kernel \( K(d) \), as defined in Eq.~\ref{eq:rbf_kernel}. Here, \( d \) denotes the distance between the NPC and the student, while \( \sigma \) is a hyperparameter that controls the scale of the kernel, determining how rapidly the influence of NPCs diminishes as their distance from the student increases. The intrinsic objective weight \( 1 - K(d) \) increases with distance, whereas the extrinsic objective weight \( K(d) \) correspondingly decreases. This weighting strategy acts as a contribution assignment mechanism based on the premise that NPCs farther from the student exert a weaker influence on the student's behavior. Consequently, distant NPCs receive a lower contribution from the extrinsic reward and are encouraged to prioritize their own driving behavior, while closer NPCs rely more on the extrinsic objective, adapting their behavior to align with the student's actions.
\begin{equation}
    K(d) = \exp\left(- \, \dfrac{d^{\,2}}{2 \cdot \sigma^{\,2}}\right)\label{eq:rbf_kernel}
\end{equation}

\subsubsection{Learning Algorithm}
At first glance, Multi-Agent Proximal Policy Optimization (MAPPO)~\cite{yu2022surprising} appears to be a suitable algorithm for training the teacher, as it extends PPO~\cite{schulman2017proximal} to a collaborative multi-agent setting. However, its reliance on a global shared reward, where all agents contribute collectively, presents a key limitation. While this approach can accelerate convergence toward a global objective, it complicates the training of individual NPC behaviors and makes it difficult to isolate each agent's contribution to the overall reward. Consequently, MAPPO is not well-suited to our proposed methodology, where NPCs must balance independent decision-making, optimizing their intrinsic goal of reaching a destination, with coordinated behavior to generate traffic dynamics at a specified difficulty level while preserving distinct contribution assignments.

Thus, we adopt Independent PPO (IPPO)~\cite{dewitt2020independentlearningneedstarcraft} as the training algorithm. Unlike standard IPPO, where each agent relies solely on its local observations, we propose a novel shared global observation processed through a graph-based architecture to compute embeddings for each agent, as discussed in section~\ref{subsection:teacher}. This approach ensures that agents are well-informed about the scene context and interactions with other agents in a differentiable manner, enabling the teacher to develop a more comprehensive understanding of the environment. Each NPC independently learns its policy and value function using individual rewards, as defined in Eq.~\ref{eq:reward}. Additionally, we incorporate policy and value function parameter sharing, allowing a single network to efficiently compute outputs for all agents within a single forward pass.

\subsection{Automatic Curriculum Algorithm}
\begin{figure}[t]
    \centering
    \includegraphics[trim={6.8cm 4.7cm 7.2cm 4.5cm},clip,width=0.95\columnwidth]{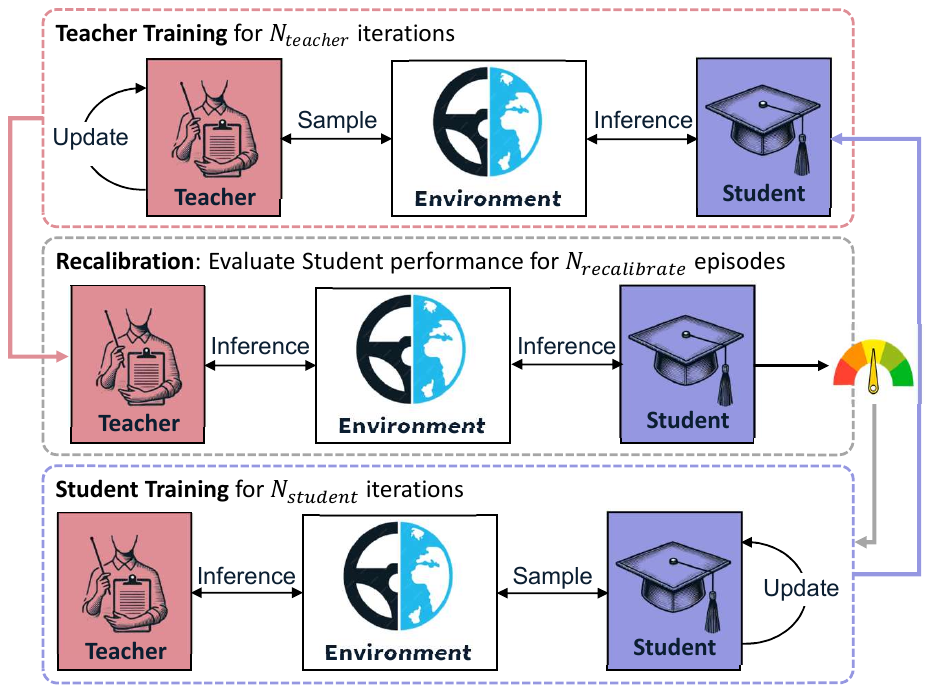}
    \caption{The proposed algorithm consists of three sequential steps. Initially, the teacher undergoes training for \( N_{\text{teacher}} \) iterations to refine its NPC behavior policy. Followed by a recalibration phase to determine the initial difficulty level of the behavior curriculum. Finally, the student driving policy is trained for \( N_{\text{student}} \) iterations under a curriculum with progressively increasing difficulty.}
    \label{fig:AutomaticCurriculum-Algo}
    \vspace{-0.6cm}
\end{figure}

We propose a novel automatic curriculum algorithm to jointly train the teacher and student RL components in a shared environment (Fig.~\ref{fig:AutomaticCurriculum-Algo}). The training process adopts an alternating Markov game~\cite{littman1996generalized}, where only one component is actively trained at a time while the other is in inference mode. This approach addresses instability issues arising from concurrent updates in a non-stationary environment, where dynamics change continuously during learning~\cite{gisslen2021adversarial}. 

The \textbf{teacher training phase} updates the NPC behavior generation policy over \( N_{\text{teacher}} \) iterations, adapting to the current student performance. During this phase, the auxiliary input \( \lambda \) is uniformly sampled from a discrete set \( \lambda_{\text{set}} \), allowing the teacher to learn behavior patterns and traffic coordination across the entire curriculum range. This process aligns the generated behaviors with the desired \( \lambda \), guided by the reward function in Eq.~\ref{eq:reward}, enabling the teacher to produce traffic at varying difficulty levels.

The \textbf{student training phase} updates the student driving policy over \( N_{\text{student}} \) iterations by navigating driving scenarios with NPC behaviors inferred by the teacher. To ensure sufficient exposure to each difficulty level, the auxiliary input \( \lambda \) remains fixed throughout each iteration. The curriculum dynamically adjusts \( \lambda \) based on the student’s success rate, defined as the proportion of episodes in which the student successfully reaches its goal in the previous iteration. When the student’s success rate exceeds a predefined threshold \( T_{\text{success}} \), the student is considered proficient at the current difficulty level, prompting an increase in challenge. Conversely, a success rate below \( T_{\text{fail}} \) indicates that the difficulty is too high, necessitating further training on easier levels before retrying the more challenging ones. If the success rate falls between these thresholds, the student training continues at the current level to reinforce learning and master this level.

To mitigate catastrophic forgetting~\cite{kirkpatrick2017overcoming} of previously mastered levels, the student has a probability \( P_{\text{old}} \) of sampling easier levels during training. The proposed self-paced mechanism ensures that the curriculum adapts to the student's performance, balancing difficulty progression while incorporating past experiences to prevent regression.

An optional \textbf{recalibration phase} is incorporated after each teacher training phase to adjust \( \lambda \), accounting for changes in traffic behavior induced by teacher policy updates. This process evaluates the student’s performance across all difficulty levels using the updated teacher policy and determines the initial \( \lambda \) for the student training phase. By adapting to the student’s performance, this approach ensures training stability in non-stationary environments while fostering adaptive learning. Its structured design generates diverse traffic behaviors, gradually increasing difficulty. 

\section{experimental Setup}
This section outlines the experimental setup, including the selected student model, the traffic scenarios, and the evaluation metrics employed to analyze the performance of the trained RL components.

\subsection{Student Description}
Our curriculum learning algorithm imposes no constraints on the student, enabling interchangeable realization and a flexible MDP setup with various observation and action spaces, network architectures, and training algorithms. In this work, the student employs a multi-modal observation comprising a frontal RGB camera with a $256 \times 256$ resolution and a LiDAR point cloud discretized into a $256 \times 256$ grid map with two height bins. Additionally, vehicle state measurements, including longitudinal and angular velocities as well as longitudinal and lateral accelerations, are incorporated into the observation space. This setup reflects the partial observability nature of real-world SDVs. To encode the input observations into a latent representation, we adopt TransFuser~\cite{chitta2022transfuser}, a transformer-based architecture designed to effectively fuse RGB and LiDAR features through cross-attention at multiple spatial scales, thereby enhancing spatial awareness. The encoded observations are subsequently utilized for estimating both the student policy and value function. The student policy is trained using PPO~\cite{schulman2017proximal}, guided by a standard driving reward ${\mathcal{R}}^{\text{driving}}_{\text{student}}$.

\subsection{Traffic Scenarios}
For the evaluation of the proposed methodology, we focus on unsignalized urban intersections, which are prevalent in urban environments and often involve safety-critical situations requiring implicit coordination between vehicles. The traffic scenarios are simulated using the CARLA simulator~\cite{dosovitskiy2017carla}. During training, NPCs are randomly positioned across three T-intersections and four four-way intersections, with assigned destinations that they navigate through the intersection. NPCs are terminated upon reaching their destination, colliding, or veering off-road. Each episode concludes when the student reaches a terminal state or exceeds a predefined time limit. To assess the capabilities of the proposed algorithm, performance is evaluated on a hold-out set consisting of one T-intersection and two four-way intersections that were not encountered during training.

\subsection{Baselines and Evaluation Metrics}
To evaluate our approach, we conducted two training experiments: one without the recalibration step (\(\text{Student}_{\,\text{CL}}\)) and one incorporating it (\(\text{Student}^{+}_{\,\text{CL}}\)). Each experiment utilized eight initial NPC agents to generate highly interactive scenarios and an auxiliary input set of nine equally spaced steps \(\lambda_{\text{set}} = \{-1, \,-0.75, \,-0.5, \,\dots, \,1\}\), ensuring steady paced curriculum progression. Additional algorithm parameters are detailed in Table~\ref{tab:params}. We compare students trained using the proposed automatic curriculum with a student trained in a baseline traffic environment that reflects rule-based NPC behavior via CARLA’s traffic manager (\(\text{Student}_{\,\text{Rule}}\)). 

To maintain consistency in evaluation, all students are trained for the same number of iterations with identical architecture and PPO parameters. The evaluation metrics include cumulative driving rewards per episode ($\mathcal{R}$) and a range of standard driving performance metrics. These metrics cover terminal statistics such as the percentage of successful episodes ($SR$), off-road deviations ($OR$), collisions ($CR$), timeouts ($TR$), and behavioral metrics like route progress ($RP$) and average driving velocity ($v$) in $m/s$.
\begin{table}[t]
\centering
\renewcommand{\arraystretch}{1.1}
\caption{Parameters of the proposed automatic behavior curriculum learning algorithm.}
\resizebox{0.95\columnwidth}{!}{
\begin{tabular}{cc|cc|cc}
\toprule
\textbf{Parameter} & \textbf{Value} & \textbf{Parameter} & \textbf{Value} & \textbf{Parameter} & \textbf{Value} \\ \midrule
$D$                & 2              & $N_{\text{teacher}}$ (Iterations)   & 10             & $T_{\text{success}}$ & 0.75           \\
$\varepsilon $     & 0.1            & $N_{\text{student}}$ (Iterations)   & 10             & $T_{\text{fail}}$    & 0.25           \\
$\sigma$           & 5.0            & $N_{\text{recalibrate}}$ (Episodes) & 100            & $P_{\text{old}}$     & 0.3 \\ \bottomrule          
\end{tabular}}
\label{tab:params}
\vspace{-0.6cm}
\end{table}
\section{Evaluation}
This section presents the results of our proposed framework. We first examine the teacher's effectiveness in generating traffic behaviors of different difficulty levels. Then, we compare the performance of students trained with our framework to that of baseline students.
\subsection{Traffic Behavior Generation}
First, we evaluate the teacher’s capability to generate varying levels of traffic difficulty by analyzing the student’s performance across five auxiliary inputs, comparing students trained with and without the recalibration step. The results presented in Table~\ref{tab:teacherAnalysis} confirm that the teacher effectively establishes a clear relationship between the auxiliary input $\lambda$ and the complexity of the generated traffic. As $\lambda$ decreases from 1 (easiest) to -1 (hardest), the student’s success rate ($SR^{\,\text{student}}$) and driving reward ($\mathcal{R}^{\,\text{student}}$) decreases, while the average NPC velocity ($v^{\,\text{NPCs}}$) increases. This trend indicates that the teacher successfully generates progressively more challenging traffic conditions. The recalibration step ($\text{Student}^{+}_{\,\text{CL}}$) enhances the smoothness and distinction of difficulty progression, ensuring that each curriculum stage accurately reflects the intended complexity. 

\begin{figure}[h]
    \centering
    
    \begin{subfigure}{0.95\columnwidth}
        \centering
        \includegraphics[width=\textwidth]
        {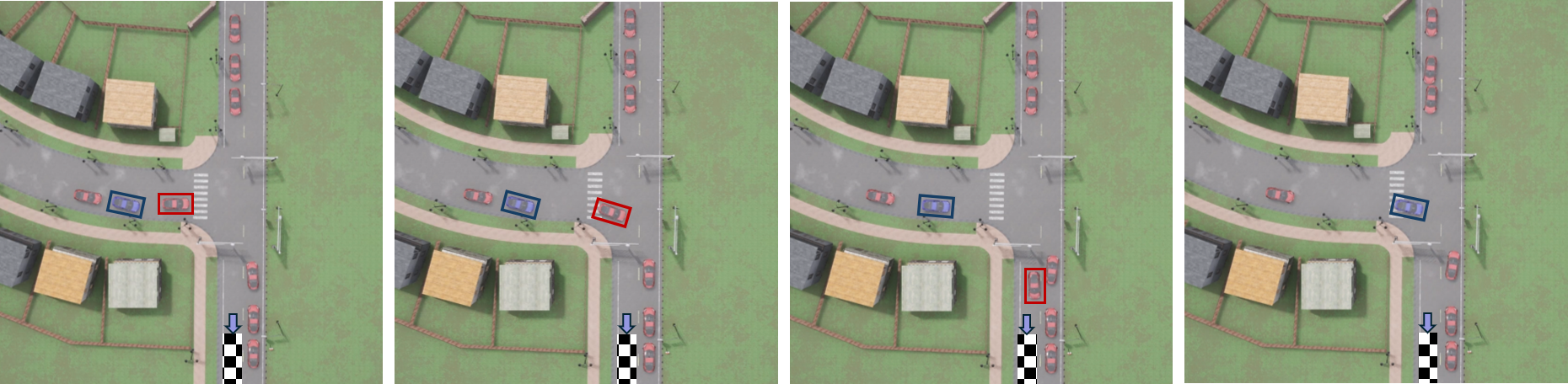}
        \vspace{-0.5cm}
        \caption{NPCs behavior at the easiest difficulty $\lambda=1$.}
    \end{subfigure} 
    \vspace{0.2cm}

    \begin{subfigure}{0.95\columnwidth}
        \centering
        \includegraphics[width=\textwidth]{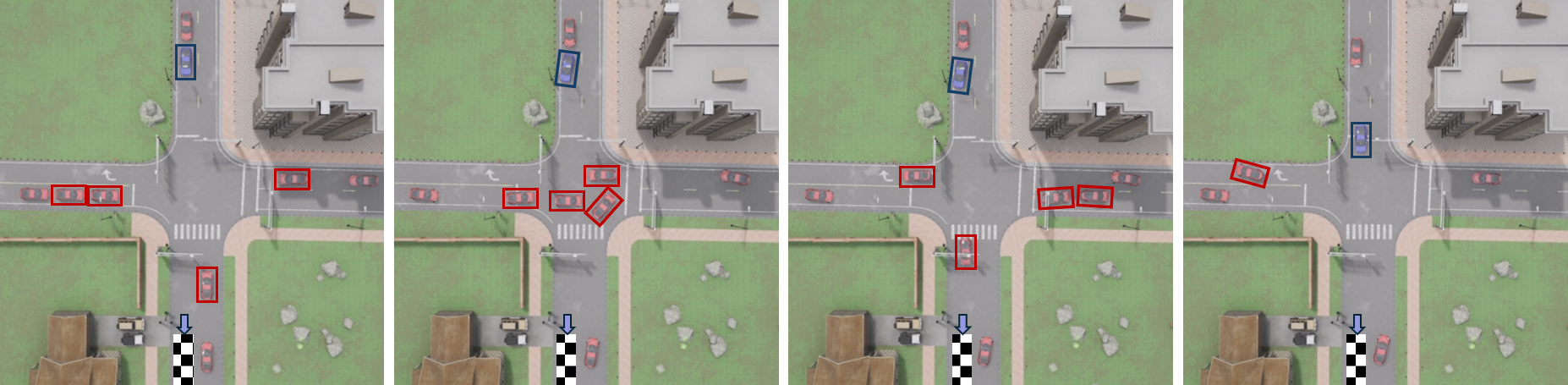}
        \vspace{-0.5cm}
        \caption{NPCs behavior at the medium difficulty $\lambda=0$.}
    \end{subfigure}%
    \vspace{0.2cm}
    
    \begin{subfigure}{0.95\columnwidth}
        \centering
        \includegraphics[width=\textwidth]{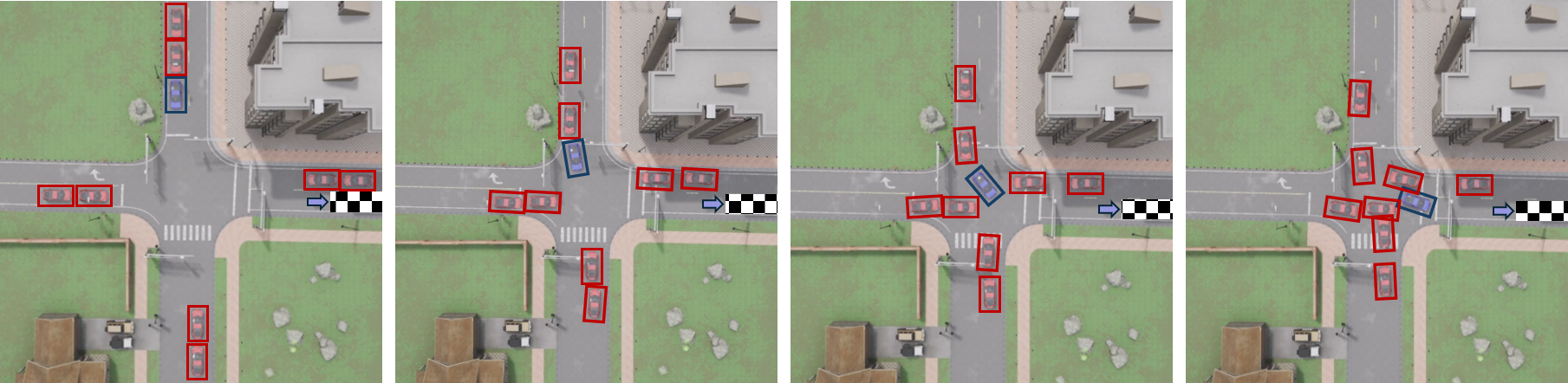}
        \vspace{-0.5cm}
        \caption{NPCs behavior at the hardest difficulty $\lambda=-1$.}
    \end{subfigure}
    \caption{Exemplary scenarios of the teacher's behavior generation across three difficulty levels, with the NPCs highlighted in red and the student highlighted in blue. Screenshots are arranged in a temporal sequence from left to right. Moving vehicles are marked with a bounding box. The black and white flag indicates the goal position of the student.}
    \label{fig:Teacher-QualitativeEval}
    \vspace{-0.6cm}
\end{figure}

Figure~\ref{fig:Teacher-QualitativeEval} illustrates qualitative scenarios at three different auxiliary input values. At $\lambda = 1$, traffic is sparse, with only the NPC directly in front of the student moving to clear the path, while others remain stationary, minimizing interactions and difficulty. At $\lambda = 0$, half of the NPCs enter the intersection while the rest wait, creating a balanced traffic flow and a moderate challenge for the student. At $\lambda = -1$, all NPCs move simultaneously into the intersection, generating dense and dynamic traffic that demands complex maneuvering. These results highlight the teacher’s ability to systematically generate increasingly complex scenarios.  
\begin{table}[t]
\centering
\caption{Evaluation of students performance across different difficulty levels $\lambda$ generated by the teacher.}
\resizebox{0.8\columnwidth}{!}{%
\begin{tabular}{c|c|c|c}
\toprule
$\lambda$ & Metric & $\text{Student}_{\,\text{CL}}$ & $\text{Student}^{+}_{\,\text{CL}}$ \\ \midrule
\cellcolor[HTML]{009901} & $SR^{\,\text{student}}$ & 0.59 & 0.65 \\
\cellcolor[HTML]{009901} & $\mathcal{R}^{\,\text{student}}$ & $0.23 \pm 1.06$ & $0.43 \pm 1.40$ \\
\multirow{-3}{*}{\cellcolor[HTML]{009901}1} & $v^{\,\text{NPCs}}$ & $0.86 \pm 0.37$ & $0.84 \pm 0.41$ \\ \midrule
\cellcolor[HTML]{9AFF99} & $SR^{\,\text{student}}$ & 0.53 & 0.57 \\
\cellcolor[HTML]{9AFF99} & $\mathcal{R}^{\,\text{student}}$ & $0.23 \pm 1.40$ & $0.23 \pm 1.32$ \\
\multirow{-3}{*}{\cellcolor[HTML]{9AFF99}0.5} & $v^{\,\text{NPCs}}$ & $0.86 \pm 0.32$ & $0.81 \pm 0.39$ \\ \midrule
\cellcolor[HTML]{FCFF2F} & $SR^{\,\text{student}}$ & 0.40 & 0.45 \\
\cellcolor[HTML]{FCFF2F} & $\mathcal{R}^{\,\text{student}}$ & $-0.43 \pm 2.02$ & $-1.13 \pm 2.80$ \\
\multirow{-3}{*}{\cellcolor[HTML]{FCFF2F}0} & $v^{\,\text{NPCs}}$ & $1.96 \pm 0.76$ & $1.71 \pm 0.33$ \\ \midrule
\cellcolor[HTML]{F8A102} & $SR^{\,\text{student}}$ & 0.44 & 0.39 \\
\cellcolor[HTML]{F8A102} & $\mathcal{R}^{\,\text{student}}$ & $-0.72 \pm 2.21$ & $-2.46 \pm 3.67$ \\
\multirow{-3}{*}{\cellcolor[HTML]{F8A102}-0.5} & $v^{\,\text{NPCs}}$ & $2.50 \pm 0.71$ & $2.39 \pm 0.37$ \\ \midrule
\cellcolor[HTML]{CB0000} & $SR^{\,\text{student}}$ & 0.45 & 0.40 \\
\cellcolor[HTML]{CB0000} & $\mathcal{R}^{\,\text{student}}$ & $-1.02 \pm 2.54$ & $-2.41 \pm 3.64$ \\
\multirow{-3}{*}{\cellcolor[HTML]{CB0000}-1} & $v^{\,\text{NPCs}}$ & $2.74 \pm 0.69$ & $2.39 \pm 0.38$ \\ \bottomrule
\end{tabular}}
\vspace{-0.6cm}
\label{tab:teacherAnalysis}
\end{table}

\subsection{Student Generalizability}
\begin{table*}[t]
\centering
\caption{Evaluation of student performance against baseline student trained in rule-based traffic on rule-based traffic and teacher-generated traffic at three different difficulty levels.}
\resizebox{0.85\textwidth}{!}{%
\begin{tabular}{c|c|ccccccc}
\toprule
Traffic Generator  & Student & $SR$ & $CR$ & $OR$ & $TR$ & $RP$            & $v$             & $\mathcal{R}$             \\ \midrule
\multirow{2}{*}{Carla Traffic Manager}  & CL & 0.72 & 0.21 & 0.07 & 0.00 & $\boldsymbol{0.72 \pm 0.32}$ & $\boldsymbol{2.21 \pm 0.80}$ & $0.16 \pm 2.72$  \\
& Rule     & 0.76 & 0.03 & 0.03 & 0.18 & $0.71 \pm 0.37$ & $1.35 \pm 0.28$ & $\boldsymbol{0.59 \pm 0.71}$  \\ \midrule

\multirow{2}{*}{Teacher ($\lambda=1$)}  & CL & 0.65 & 0.31 & 0.00 & 0.04 & $\boldsymbol{0.66 \pm 0.36}$ & $\boldsymbol{1.81 \pm 1.14}$ & $\boldsymbol{0.43 \pm 1.40}$  \\
& Rule     & 0.43 & 0.38 & 0.01 & 0.18 & $0.47 \pm 0.43$ & $0.81 \pm 0.92$ & $-0.15 \pm 1.24$ \\ \midrule

\multirow{2}{*}{Teacher ($\lambda=0$)}  & CL & 0.45 & 0.53 & 0.02 & 0.00 & $\boldsymbol{0.52 \pm 0.40}$ & $\boldsymbol{1.73 \pm 1.11}$ & $\boldsymbol{-1.13 \pm 2.80}$ \\
& Rule     & 0.47 & 0.53 & 0.00 & 0.00 & $0.45 \pm 0.45$ & $0.86 \pm 0.92$ & $-1.72 \pm 2.81$ \\ \midrule

\multirow{2}{*}{Teacher ($\lambda=-1$)} & CL & 0.40 & 0.58 & 0.02 & 0.00 & $\boldsymbol{0.45 \pm 0.41}$ & $\boldsymbol{1.63 \pm 1.06}$ & $\boldsymbol{-2.41 \pm 3.64}$ \\
 & Rule     & 0.41 & 0.59 & 0.00 & 0.00 & $0.40 \pm 0.45$ & $0.82 \pm 0.94$ & $-2.62 \pm 3.17$ \\ \bottomrule
\end{tabular}}
\label{tab:robustnessComparison}
\end{table*}
We compare students trained using the proposed curriculum learning algorithm against baseline students trained in rule-based traffic. The student performance is assessed in scenarios where NPC behaviors are governed by the rule-based CARLA traffic manager, as well as across three difficulty levels generated by a learned teacher within the proposed framework. The results, demonstrated in Table~\ref{tab:robustnessComparison}, indicate that students trained with an automatic curriculum generalize more effectively, achieving safer navigation in rule-based traffic while exhibiting higher route progress and average velocity. Additionally, students trained with curriculum learning outperform the baseline across all three teacher-generated difficulty levels, demonstrating significant improvements in route progress, average driving velocity, and overall driving rewards. 

\begin{figure*}[h]
    \centering
    \begin{subfigure}[t]{0.23\linewidth}
    \centering
    \includegraphics[trim={0.2cm 0.2cm 0.2cm 1.5cm},clip,width=\linewidth]{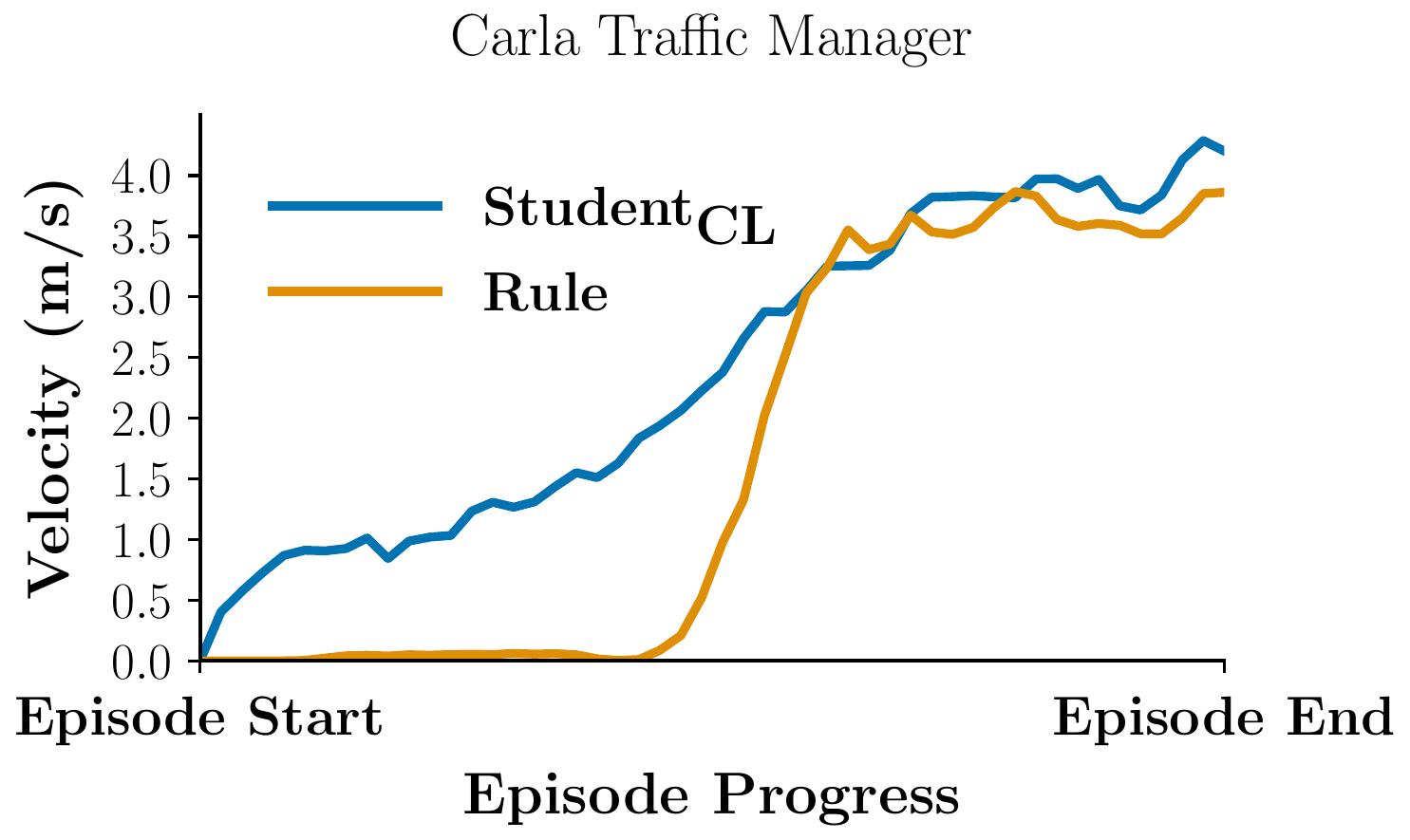}
    \caption{Velocity profiles at CARLA traffic manager (\textit{rule-based}).}
    \end{subfigure}
    \hfill
    \begin{subfigure}[t]{0.24\linewidth}
    \centering
    \includegraphics[trim={0.2cm 0.2cm 0.2cm 1.5cm},clip,width=\linewidth]{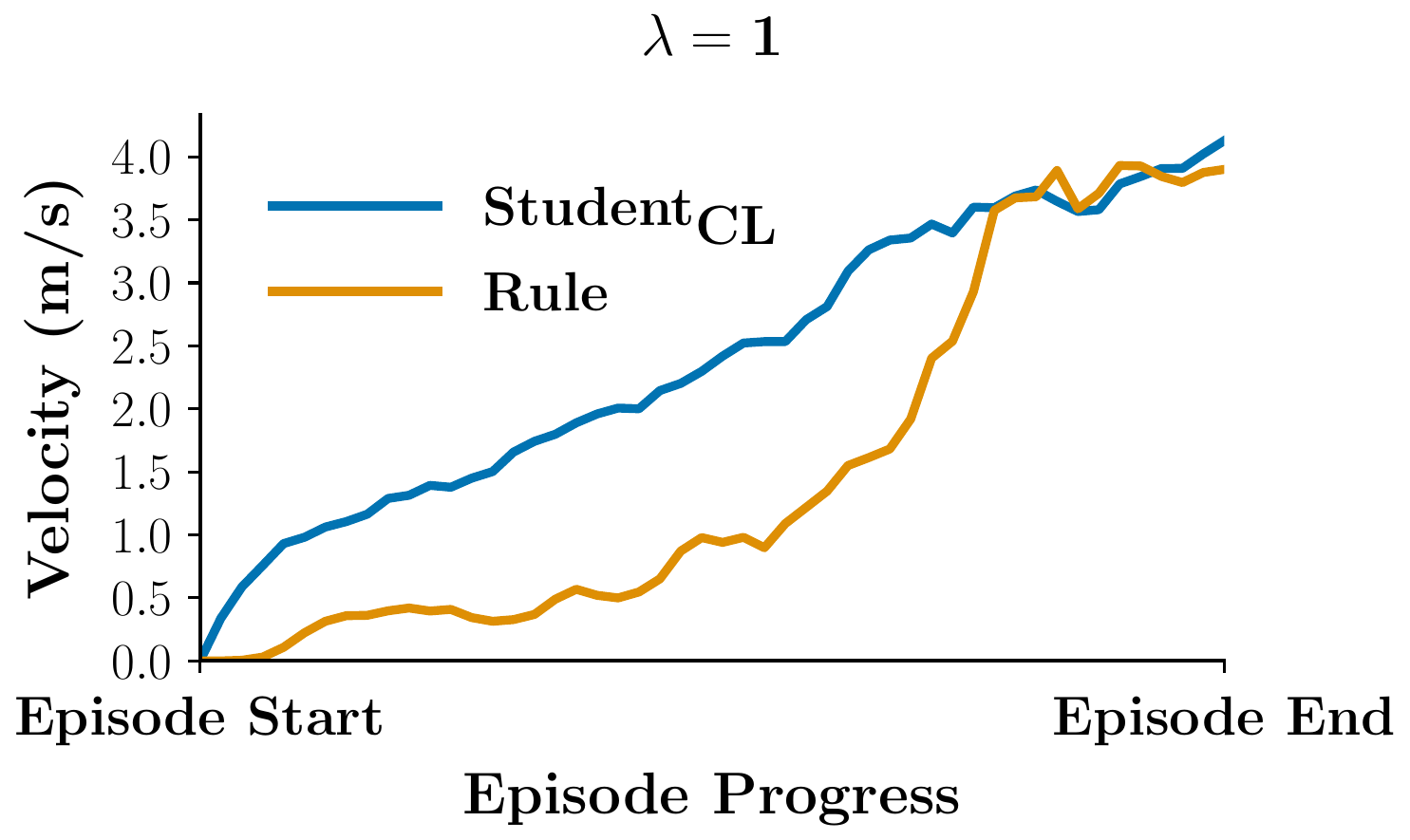}
    \caption{Velocity profiles at the teacher's easiest difficulty $\lambda=1$.}
    \end{subfigure}
    \hfill
    \begin{subfigure}[t]{0.24\linewidth}
    \centering
    \includegraphics[trim={0.2cm 0.2cm 0.2cm 1.5cm},clip,width=\linewidth]{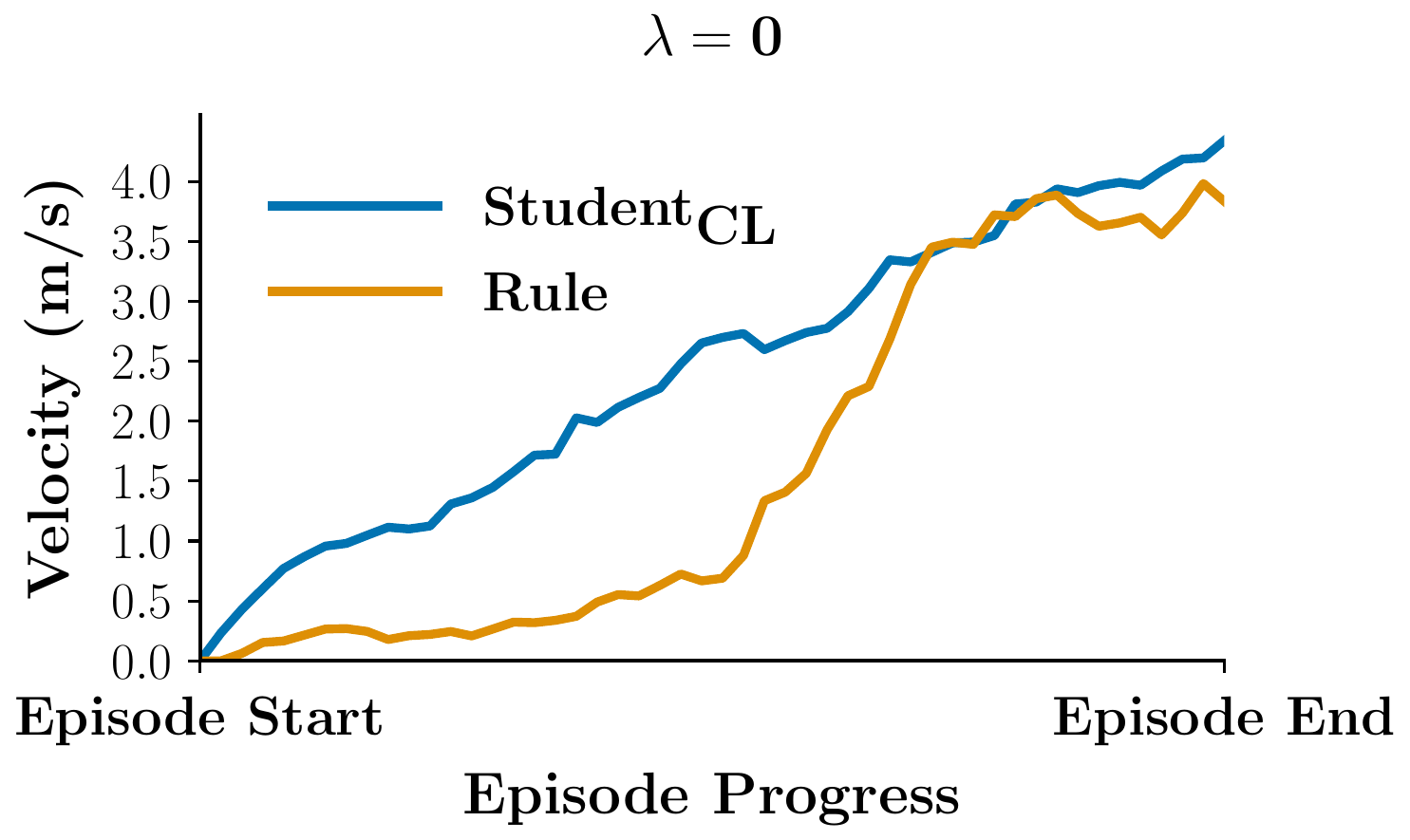}
    \caption{Velocity profiles at the teacher's medium difficulty $\lambda=0$.}
    \end{subfigure}
    \hfill
    \begin{subfigure}[t]{0.24\linewidth}
    \centering
    \includegraphics[trim={0.2cm 0.2cm 0.2cm 1.5cm},clip,width=\linewidth]{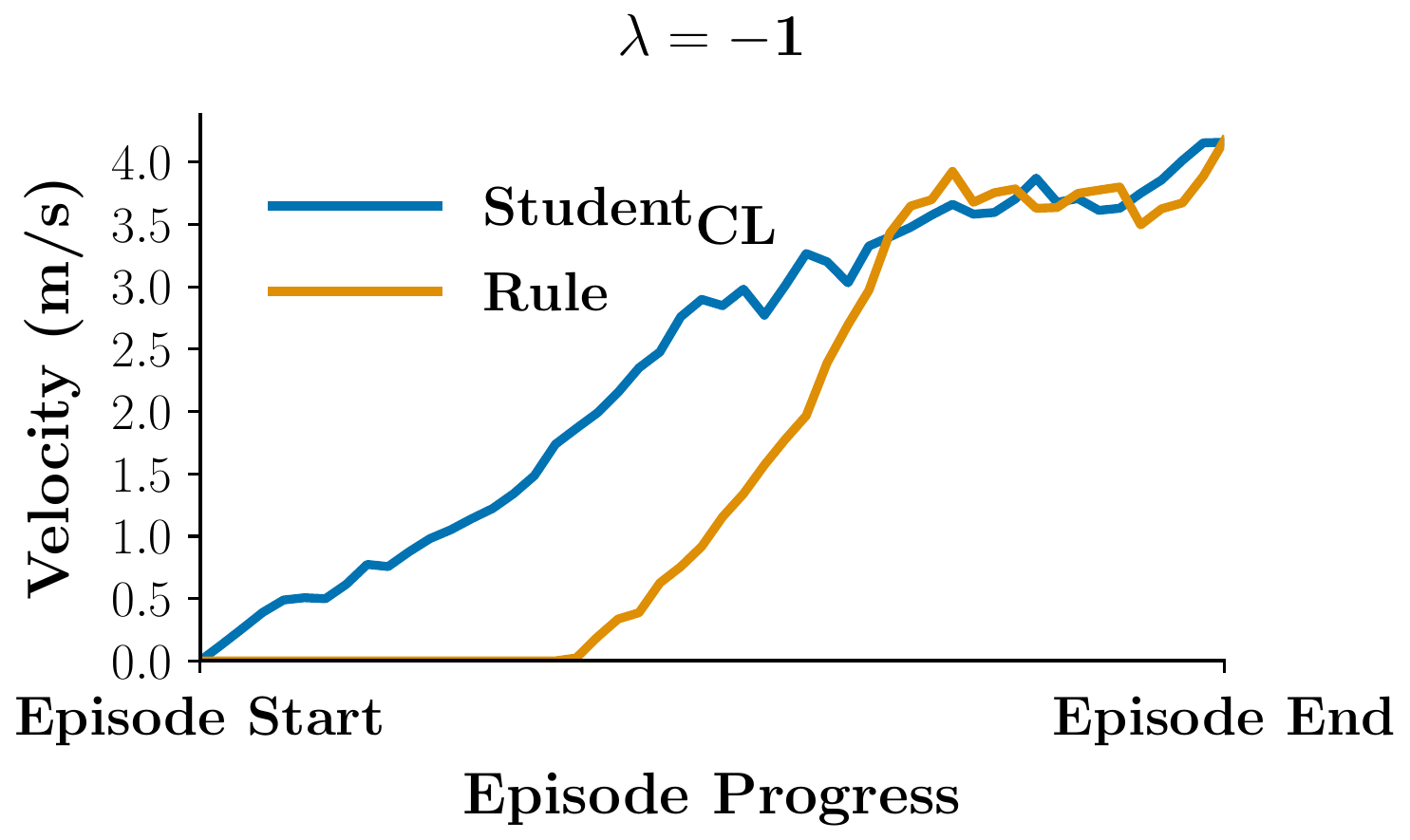}
    \caption{Velocity profiles at the teacher's hardest difficulty $\lambda=-1$.}
    \end{subfigure}
    \caption{Comparison of student velocity profiles across different traffic conditions. Each plot shows two students: one trained with curriculum learning (CL) and one with rule-based traffic. The rule-based student adopts an exploitative policy, often waiting passively for NPCs to clear the intersection. Conversely, the CL student navigates traffic more assertively and maintains smoother, balanced velocity.}
    \vspace{-0.6cm}
    \label{fig:velocity_profile}
\end{figure*}

Further qualitative analysis of the learned student policies exposes a critical flaw in the baseline student. While it occasionally achieves higher success rates, this is a direct consequence of an exploitative policy that merely waits for all NPCs to clear the intersection before proceeding, as investigated in Fig~\ref{fig:velocity_profile}. In contrast to this unrealistic behavior, students trained with curriculum learning exhibit a more adaptive driving behavior. They proactively engage with traffic, demonstrating realistic and intuitive decision-making to safely navigate intersections. This enhanced adaptability is further validated by their consistently higher average velocity and route progress, underscoring the superiority of curriculum learning in fostering interactive driving policies.
\section{Conclusion}
This work proposes a student-teacher framework to improve SDV robustness in dynamically generated traffic using MARL automatic curriculum learning. The teacher adapts NPCs' behavior, adjusting difficulty levels based on student performance. Evaluations show that the teacher successfully generates diverse traffic behaviors across difficulty levels, enabling students trained with the dynamic curriculum to outperform those trained on rule-based traffic. 
Future research will refine the teacher’s reward to enhance traffic coordination by incorporating the surrounding NPCs' rewards and expand the curriculum to include cyclists and pedestrians for broader agent interaction.
\section*{ACKNOWLEDGMENT}
The research leading to these results is funded by the German Federal Ministry for Economic Affairs and Energy within the project “Safe AI Engineering – Sicherheitsargumentation befähigendes AI Engineering über den gesamten Lebenszyklus einer KI-Funktion". The authors would like to thank the consortium for the successful cooperation.

{
    \bibliographystyle{IEEEtran}
    \bibliography{references}
}

\end{document}